\documentclass{article} % For LaTeX2e
\usepackage[final]{colm2025_conference}

\usepackage{microtype}
\usepackage{hyperref}
\usepackage{url}
\usepackage{booktabs}

\usepackage{adjustbox}
\usepackage{graphicx}
\usepackage{colortbl}
\usepackage{lineno}
\usepackage{multirow}
\usepackage{graphicx}
\usepackage{subcaption}
\usepackage{dsfont,bbm}
\usepackage{xspace}
\usepackage{xcolor}
\usepackage{soul}
\definecolor{pptyellow}{RGB}{254,255,84}
\definecolor{green}{RGB}{233,219,169}
\definecolor{bright}{RGB}{253,247,167}

\newcommand{\say}[1]{``#1''}

\newcommand{\eg}{\emph{e.g.}}

\newcommand{\BenchFull}{Multimodal Academic Cover benchmark\xspace}
\newcommand{\BenchFullShort}{MAC\xspace}
\newcommand{\latestBench}{MAC-2025\xspace}

\newcommand{\ourMethod}{Description and Deduction\xspace}
\newcommand{\ourMethodShort}{DAD\xspace}

\newcommand{\ttoi}{Text2Image\xspace}
\newcommand{\itot}{Image2Text\xspace}

\usepackage{amsmath,amsfonts,bm}

\def\figref#1{Figure~\ref{#1}}
\def\Figref#1{Figure~\ref{#1}}
\def\tabref#1{Table~\ref{#1}}
\def\Tabref#1{Table~\ref{#1}}

\def\secref#1{section~\ref{#1}}
\def\eqref#1{equation~\ref{#1}}
\def\1{\bm{1}}

\DeclareMathAlphabet{\mathsfit}{\encodingdefault}{\sfdefault}{m}{sl}
\SetMathAlphabet{\mathsfit}{bold}{\encodingdefault}{\sfdefault}{bx}{n}

\definecolor{darkblue}{rgb}{0, 0, 0.5}
\hypersetup{colorlinks=true, citecolor=darkblue, linkcolor=darkblue, urlcolor=darkblue}

\title{\BenchFullShort: A Live Benchmark for Multimodal Large Language Models in Scientific Understanding}

\author{
Mohan Jiang$^{1,2}$\thanks{~Equal contribution. $^\dag$ Corresponding author: \href{mailto:dequanwang@sjtu.edu.cn}{dequanwang@sjtu.edu.cn}.} ,
Jin Gao$^{1*}$,
Jiahao Zhan$^3$,
Dequan Wang$^{1,2\dag}$ \\
$^1$Shanghai Jiao Tong University \quad
$^2$Shanghai Innovation Institute \quad
$^3$Fudan University \\
}

\begin{document}

\ifcolmsubmission
\linenumbers
\fi

\maketitle

\begin{abstract}
As multimodal large language models (MLLMs) become increasingly capable, fixed benchmarks are gradually  losing their effectiveness in evaluating high-level scientific understanding.
In this paper, we introduce the \BenchFull (\BenchFullShort), a live benchmark that could continuously evolve with scientific advancement and model progress.
\BenchFullShort leverages over 25,000 image-text pairs sourced from issues of top-tier scientific journals such as Nature, Science, and Cell, challenging MLLMs to reason across abstract visual and textual scientific content.
Experiments on our most recent yearly snapshot, \latestBench, reveal that while MLLMs demonstrate strong perceptual abilities, their cross-modal scientific reasoning remains limited.
To bridge this gap, we propose \ourMethodShort, a lightweight inference-time approach that enhances MLLM by extending visual features with language space reasoning, achieving performance improvements of up to 11\%.
Finally, we highlight the live nature of \BenchFullShort through experiments on updating journal covers and models for curation, illustrating its potential to remain aligned with the frontier of human knowledge. 
We release our benchmark at \href{https://github.com/mhjiang0408/MAC_Bench}{https://github.com/mhjiang0408/MAC\_Bench}.
\end{abstract}

\section{Introduction}

With extensive research driving into developing multimodal large language models (MLLMs) for autonomous scientific research based on their exceptional capabilities in complex domains, evaluating MLLMs for scientific understanding remains a long-standing and evolving challenge. In earlier stages, benchmarks such as MMMU~\citep{yue2024mmmu} served as demanding tests of multimodal reasoning, particularly in scientific domains. However, as MLLMs grow increasingly capable, such benchmarks show signs of saturation. For example, Gemini-2.5-pro~\citep{gemini2.5pro} recently achieved 81.7\% accuracy in a pass@1 setting on MMMU, suggesting that even with comprehensive curation, static benchmarks may lose their guiding influence over time.

This phenomenon highlights the need for a \emph{live} benchmark—one that evolves alongside model progress, rather than remaining static.
Recent efforts like LiveBench~\citep{white2024livebench} aim to address this by regularly releasing new questions sourced from dynamic content such as arXiv papers, news articles, and IMDb synopses.
However, LiveBench remains unimodal, focusing solely on language, and its reliance on rapidly changing content from the Internet raises concerns about data quality in scientific understanding.

\begin{figure}[htb]
    % \flushleft
    \centering
    \includegraphics[width=\linewidth]{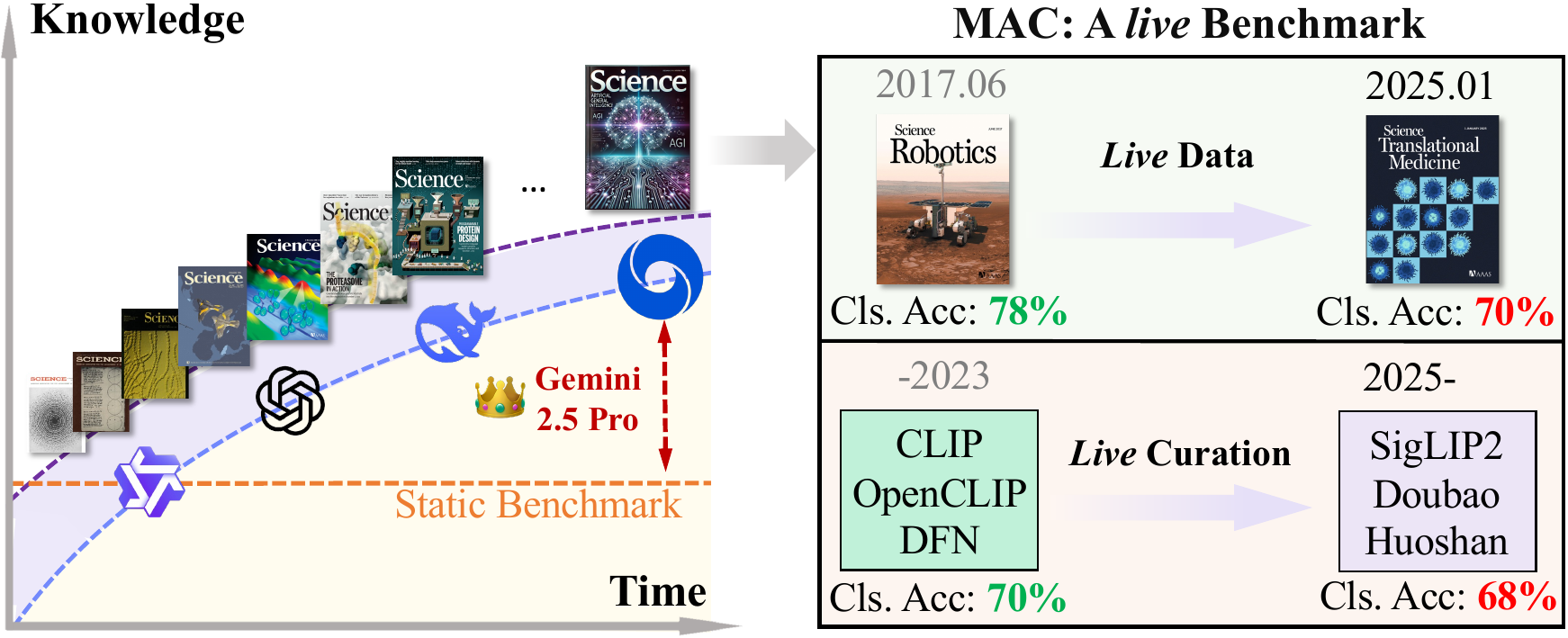}
    \caption{\textbf{\BenchFullShort is a \emph{live} benchmark to evaluate the scientific understanding of MLLMs.} By continuously incorporating the latest scientific discoveries and curating with the latest embedding models, our \emph{live} benchmark transcends the limitations of static benchmarks that face progressive performance saturation by developing MLLMs.
    Best viewed in color.
    % Cls. Acc. means the classification accuracy.
}
    \label{fig:cover_teaser}
\end{figure}

Leading scientific journals, carefully curated by experts, show great potential as they publish new issues weekly or monthly.
Each issue contains a journal cover and a corresponding cover story, depicting the same scientific topic from vision and language modalities.
These image-text pairs encapsulate complex scientific concepts through abstract multimodal representation.
As cutting-edge discoveries, they often fall outside existing model pretraining corpora, making them ideal testbeds for probing scientific understanding in MLLMs.

Motivated by this, we introduce the \BenchFull (\BenchFullShort), a \emph{live} benchmark built from \emph{live} scientific journals and curated through our \emph{live} curation mechanism to test MLLMs' scientific visual understanding and cross-modal concept alignment capabilities, shown in \figref{fig:cover_teaser}.
Each question presents a four-way classification task: given a cover image, select the corresponding cover story (\itot), or given a story, select the matching cover (\ttoi).
Distractors are selected by embedding similarity across multiple models, ensuring the benchmark remains challenging even for the latest MLLMs.

We present \latestBench, the most recent yearly snapshot of \BenchFullShort, constructed from over 2,000 cover image-story pairs drawn from premier journals such as Nature, Science, and Cell, covering issues from January 2024 to February 2025. We evaluate seven leading MLLMs on both \itot and \ttoi tasks, revealing that while they exhibit strong visual recognition capabilities,
such performance heavily relies on the text on cover images.
% they struggle with higher-level cross-modal scientific reasoning.
% For example, models can identify prescriptions and pills in images but fail to connect these elements with core concepts like \say{therapy resistance}.

% To help model deal with the complex visual information in \latestBench, we introduce \ourMethod (\ourMethodShort), a simple yet effective inference-time approach, which can allow MLLMs to not just “see,” but also “understand” and “think.”
% \ourMethodShort builds a two-stage cross-modal reasoning pipeline that first extracts multi-granularity visual features using an MLLM, and then applies a language model to perform high-level reasoning.
% This hybrid approach yields consistent gains of 1\%–5\% across baselines, with ernie-4.5-8k achieving an 11\% improvement, showcasing its effectiveness in giving more intellect to MLLMs.

To better handle the challenging visual information in \latestBench, we introduce \ourMethod (\ourMethodShort), a simple yet effective inference-time approach that enables MLLMs not only to perceive visual input, but also to understand and reason across modalities. \ourMethodShort builds a two-stage cross-modal reasoning pipeline that first extracts multi-granularity visual features using an MLLM, and then applies a language model to perform high-level reasoning. This hybrid approach yields consistent gains of 1\%–5\% across baselines, with Ernie-4.5-8k~\citep{ernie} achieving an 11\% improvement, demonstrating its ability to equip MLLMs with greater intellectual capacity in scientific visual understanding.

Beyond the snapshot, we explore the \emph{live} nature of \BenchFullShort from two perspectives (\figref{fig:cover_teaser}), \emph{live} data and \emph{live} data curation. To assess data evolution, we collect a large-scale dataset comprising over 25,000 image-text pairs from past and current journal issues.
Our analysis shows that MLLMs perform better on older issues, underscoring the increasing complexity and novelty of recent scientific content.
For benchmark construction, we regenerate \latestBench using the latest embedding models released in 2025 (concurrent to our work), which yield harder distractors and further degrade model accuracy, demonstrating our benchmark’s adaptability to advances in representation learning.

Our contributions are shown as follows:
\begin{itemize}
    \item We introduce \BenchFullShort, a continuously updating benchmark for evaluating multimodal scientific understanding in MLLMs. Based on live scientific journals, our benchmark is built using a live mechanism that adapts to model progress.

    \item We study the latest yearly snapshot of \BenchFullShort, \latestBench, drawn from over 2,000 curated journal issues, and provide a thorough evaluation of seven advanced MLLMs on both \itot and \ttoi tasks.
    
    \item We propose an inference-time approach, \ourMethod (\ourMethodShort). It significantly enhances MLLMs’ scientific concept reasoning by bridging cross-modal information between MLLMs and a reasoning language model.

    \item We investigate the live attribute of our \BenchFullShort through temporal data analysis and adaptive benchmark construction, showing the necessity of growing scientific journals and evolving construction using embedding models.

\end{itemize}

\section{Related Work}
\paragraph{Multimodal Large Language Model}

Since the foundational work of~\cite{radford2021learning} on joint image-text representations, multimodal large language models (MLLMs) have rapidly advanced, leading to a series of notable models such as~\cite{zeng2022glm},~\cite{driess2023palm},~\cite{achiam2023gpt},~\cite{liu2023visual}, and~\cite{glm2024chatglm}. Recent progress has shown their strong potential across a wide range of tasks and applications~\citep{huo2024mmneuron, xi2025rise}. Proprietary models like GPT-4o~\citep{hurst2024gpt} and Claude3.5~\citep{claude35} have achieved outstanding results on benchmarks~\citep{song2024milebench, wang2024charxiv}, while open-source models such as LLaVA-NeXT~\citep{liu2024llavanext}, DeepSeek-VL~\citep{lu2024deepseek}, InternVL 2.5~\citep{chen2024expanding} and Qwen2.5-VL~\citep{bai2025qwen2} leverage advanced projection techniques to integrate visual and textual features for efficient multimodal understanding. These developments underscore the growing impact of MLLMs in both research and real-world applications. In this context, we introduce \BenchFull~(\BenchFullShort)—a benchmark specifically designed to assess MLLMs’ ability to comprehend implicit scientific concepts in cover images.

\paragraph{MLLM benchmark}

With the rapid advancement of MLLMs, many benchmarks have emerged, highlighting the importance of evaluating both perception-understanding and cognition-reasoning capabilities.~\cite{liu2024mmbench} focused on basic perception tasks like object localization via multiple-choice questions, while~\cite{wang2024muirbench}, \cite{meng2024mmiu}, and \cite{kil2024compbench}
assessed multi-image understanding and comparative reasoning through complex visual tasks. \cite{lu2023mathvista} and \cite{cao2024visual} examined logical reasoning using abstract visual questions and mathematical problems. Cross-domain knowledge was evaluated by \cite{lu2022learn}, \cite{zhang2023m3exam}, and \cite{yue2024mmmu} using multidisciplinary questions across educational levels. However, most existing benchmarks are static and struggle to remain challenging as MLLMs rapidly evolve~\citep{bai2025qwen2,lu2024deepseek,chen2024expanding}, lacking mechanisms for regular updates or scalable expansion.
In contrast, our \BenchFullShort offers a sustainable evaluation framework that maintains appropriate difficulty while accurately measuring scientific comprehension, enabling continuous assessment of MLLMs' scientific visual understanding and cross-modal concept alignment
capabilities.

\paragraph{Scientific Figure Question-Answering Benchmark}

Evaluating MLLMs' ability to understand scientific figures is critical for assessing their deeper reasoning skills, where \cite{li2025science} released concurrent to our work reinforces this importance by incorporating scientific knowledge into generative models through Science-T2I and SciScore. Early benchmarks~\citep{kahou2017figureqa, chen2020figure} focused on synthetic chart-based VQA datasets (\eg, line and bar graphs), emphasizing visual parsing over scientific understanding. Later efforts~\citep{masry2022chartqa, methani2020plotqa, li2023scigraphqa, hu2024mplug, li2024multimodal, pramanick2024spiqa,li2024mmsci} introduced more realistic figures, from real-world scenarios or arXiv papers across disciplines. However, these datasets typically rely on figure-caption pairs, where captions may omit key scientific insights encoded in the visual data, limiting the evaluation to surface-level format comprehension. Moreover, the quality and completeness of captions can vary significantly, introducing noise and inconsistency. \latestBench addresses this gap by using covers and cover stories from 188 journals across four major publishers. Crafted by professional artists and editors, these pairs exhibit strong semantic alignment, embedding rich scientific meaning in both text and imagery. This enables a more rigorous test of MLLMs’ ability to interpret high-level scientific visual concepts.
\section{Benchmark}

\subsection{Dataset Collection and Structure}

To support a comprehensive benchmark, we first present a large-scale repository of 188 academic journals, including main and subsidiary titles from leading scientific journal presses from their inception to February 2025. Each issue is represented as a tuple of (1) \textit{Cover Image}, which displays the front cover of the issue, and (2) \textit{Cover Story} that explains the scientific concept behind the cover and introduces the featured article. To keep the benchmark aligned with the cutting edge of scientific knowledge and challenging for current MLLMs, we construct \latestBench, the latest yearly snapshot of \BenchFullShort from January 2024 to February 2025, which serves as the primary focus of our empirical analysis.

The journal collection in \BenchFullShort is sourced from the official websites of four renowned journal series, which are briefly introduced below.

% Cell\cite{cell}及其子刊：\Dataset 收录了 42 本隶属于 Cell 系列的子刊，这些期刊主要关注分子与细胞生物学，总计 7619 期。
% Nature\cite{nature}及其子刊：该部分 \Dataset 涵盖了来自 Nature 系列的 67 本子刊。作为以多学科特点著称的期刊，Nature 系列覆盖了包括材料、药物、能源、基因等在内的广泛学科领域。
% Science\cite{science}及其子刊：在这一部分 \Dataset 收集了 Science 系列下的全部六本期刊，总计 2792 期，重点展示了物理、医学、工程等多个学科领域，体现了跨学科交叉研究的特性。
% American Chemical Society (ACS)\cite{acs} 旗下刊物：该部分 \Dataset 包含了 ACS 出版社旗下的 74 本期刊，共计 8007 期。ACS 期刊不仅关注传统化学领域（如有机化学），还涵盖材料科学、化学工程等交叉领域，体现出较强的学科融合特征。

\begin{itemize}
    \item \textbf{Cell~\citep{cell} and its sub-journals}: \BenchFullShort includes 42 sub-journals from the Cell series, which primarily focus on molecular and cell biology, comprising a total of 7,619 issues;
    \item \textbf{Nature~\citep{nature} and sub-journals}: This segment includes 67 journals from the Nature series, known for its multidisciplinary coverage across fields such as materials, pharmaceuticals, energy, and genetics.
    \item \textbf{Science~\citep{science} and its sub-journals}: \BenchFullShort includes all six journals in the Science family, totaling 2,792 issues, covering interdisciplinary research across physics, medicine, and engineering.;
    \item \textbf{American Chemical Society (ACS) (\cite{acs}) Publications}: This segment includes 73 journals with 8,004 issues, covering core chemistry fields and interdisciplinary areas such as materials science and chemical engineering.
\end{itemize}
% \begin{figure}[htb]
%     \centering
%     \includegraphics[width=1\linewidth]{Figure/dataset.pdf}
%     \caption{\textbf{Dataset collection pipeline.} Our MAC-2025 covers 188 authoritative scientific journals since 2024, with a total of 2,287 cover-story pairs spanning 34 fields. Each cover image-story pair has a semantic correspondence embedded in the long text story.}
%     % 数据集收集pipeline。我们的MAC-2025涵盖了自2024年以来的188个权威科学期刊，总计2287个封面-故事对，覆盖了34个领域。每一个封面图片-故事对都拥有隐含在长文本故事中的语义对应。
%     \label{fig:dataset}
% \end{figure}

% \geometry{a4paper, margin=1in}

\begin{figure}[htb]
    \centering
    \includegraphics[width=\linewidth]{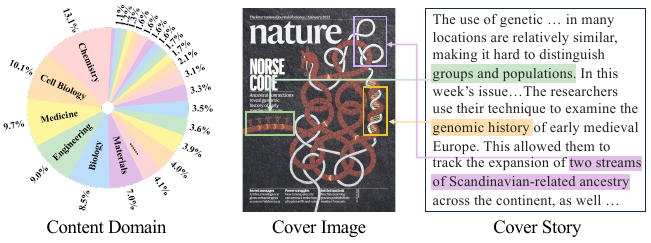}
    \caption{
    \textbf{Our \BenchFullShort spans diverse scientific disciplines, offering comprehensive visual and textual scientific materials.}
    \textit{Left} presents the distribution of disciplinary categories, while \textit{right} demonstrates representative examples of visual-textual scientific concept alignments from our dataset.
    Best viewed in color.
    }
    \label{fig:dataset}
\end{figure}

% 这些期刊以卓越的研究信誉著称，其封面设计精美且与期刊文章内容紧密相关。此外，每个期刊均配有专业编辑撰写的封面故事，内容翔实，蕴含丰富的科学概念。我们通过Open Access从网络获取这些期刊的Cover和Story并清洗掉其中Cover Image完全相同的样本。由于数据集涵盖的科学内容处于各自领域的前沿，且通常采用复杂语言表达，因此对多模态大模型（MLLMs）来说，理解这些图片和文字具有较高的挑战性。
% The journals included in \Dataset are renowned for their scientific authority and high-quality editorial standards. Their cover designs are not only visually compelling but also closely aligned with the featured articles, while each issue is accompanied by a professionally written cover story offering detailed explanations and rich scientific concepts, as illustrated in \figref{fig:dataset}. We collect these cover-story pairs from open-access sources and remove duplicates with identical cover images. As the content spans the forefront of various scientific domains and is often conveyed in complex, discipline-specific language, interpreting this information presents a significant challenge for multimodal large language models (MLLMs).
\BenchFullShort comprises cover-story pairs from authoritative scientific journals known for their high editorial standards and visually compelling, content-aligned covers. Each issue includes a professionally written cover story conveying rich scientific concepts (\figref{fig:dataset}), often expressed in complex, domain-specific language, posing challenges for MLLMs. We collect these pairs from open-access sources and remove duplicates with identical images.

% The knowledge content of static benchmarks is quickly absorbed by MLLMs during training, reducing their long-term evaluative value. To keep our benchmark challenging and relevant, we focus on recent scientific breakthroughs. Specifically, we curate a latest snapshot of \Dataset as the dataset for \latestBench, containing 2,287 samples from discoveries published between January 2024 and February 2025. These samples represent cutting-edge knowledge rarely seen in training corpora, enabling more accurate evaluation of MLLMs’ scientific understanding. To address the limitations of static benchmarks, we adopt an annual release schedule aligned with journal publication cycles, continuously incorporating new content. This dynamic update mechanism supports ongoing tracking of scientific and model evolution. Building on this, MAC-2025 offers a sustainable, evolving benchmark that keeps pace with MLLM progress while maintaining rigorous standards for visual and scientific comprehension.
To address the rapid saturation of existing benchmarks, we curate a recent snapshot—\latestBench—containing 2,287 samples from discoveries published between January 2024 and February 2025. These cutting-edge samples, rarely seen in training data, allow for a more accurate and forward-looking evaluation of MLLMs’ scientific understanding. To overcome the limitations of static benchmarks, we adopt an annual release schedule with each year's latest snapshot released every March, aligned with journal publication cycles, enabling dynamic updates that reflect both scientific and model evolution. MAC-2025 thus serves as a sustainable, evolving benchmark that maintains challenging standards while keeping pace with MLLM progress.

\subsection{\latestBench}
\label{subsec:benchmark}
\begin{figure}[htb]
    \centering
    \includegraphics[width=\linewidth]{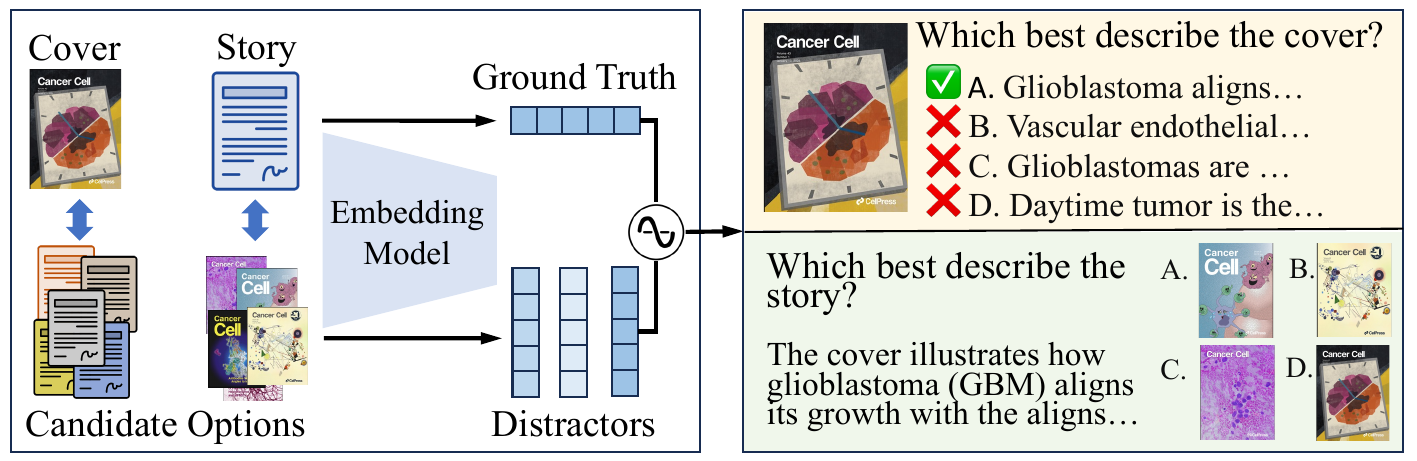}
    \caption{\textbf{Construction pipeline and benchmark examples.} Our benchmark incorporates four distinct evaluation tracks, leveraging two categories of semantic similarity comparisons through embedding model representations and bidirectional understanding tasks encompassing both image-to-text and text-to-image understanding.
    Best viewed in color.
    }
    \label{fig:method}
\end{figure}

We design two modality-specific tasks shown in \figref{fig:method} based on the cover images and stories to assess the performance of MLLMs in processing multimodal information and achieving target objectives. Our tasks are presented in the form of classification questions.

\paragraph{Bidirectional Tasks} We evaluate the model through two bidirectional multimodal tasks, assessing its ability to perceive visual elements in images and to understand the scientific concepts embedded in both text and visuals, and thereby examine its performance in cross-modal semantic alignment and reasoning.
The \itot task is selecting cover stories given cover images.
We first present MLLMs with a journal cover and ask them to select the most relevant cover story from four options. 
The \ttoi task is selecting cover images given cover stories.
% We require the MLLMs to select a cover image from four candidate covers, matching the provided cover story.
MLLMs are required to identify the cover image, among four candidates, that corresponds most accurately to the provided cover story.

\paragraph{Curation of Distractors}
\label{para:distractors}

For Image2Text and Text2Image classification tasks, distractors are drawn from cover stories within the same journal to ensure similar knowledge domains. We introduce two selection methods to maximize confusion for the models from two domains, shown in the left part of \figref{fig:method}.
The information domain (info domain) means selecting distractors based on a higher similarity to the information provided in the task.
The option domain (option domain) means selecting distractors based on a higher similarity to the option that serves as the ground truth in the task.

To build similarity metrics for the info and option domains~\citep{li2025science}, we select three embedding models per modality and compute cosine similarities.
For text-only comparisons (Image2Text option domain), we use three SentenceTransformer models~\citep{reimers-2020-multilingual-sentence-bert}, multi-qa-mpnet-base-dot-v1, all-mpnet-base-v2, and paraphrase-MiniLM-L6-v2; for image or cross-modal comparisons (Text2Image full task and Image2Text info domain), we use three multimodal visual encoders, clip-vit-large-patch14-336~\citep{radford2021learning}, ViT-g-14~\citep{ilharco_gabriel_2021_5143773}, and ViT-H-14-quickgelu~\citep{fang2023data}. 
These models, trained on diverse datasets with varying scales, provide complementary strengths. 
For each classification question, we select the top-ranked distractor from each model and choose the final one based on the highest average rank when overlaps occur.

\paragraph{Self-Validation Experiment}

While we use embedding similarity to select challenging distractors, this raises two concerns: whether embedding models can solve our bidirectional matching tasks using similarity alone, and whether the scores truly reflect semantic alignment. To explore this, we conduct self-verification experiments across both modalities (\tabref{tab:embedding-self-proof}). Results show that although embedding models achieve under 40\% accuracy—revealing their limitations in cross-modal understanding—their self-rankings fall within the top 25\%, suggesting the selected distractors are semantically close and effectively challenging. Additional analysis is discussed in \secref{sec:self-validation} and \secref{subsec:error}.

\subsection{\ourMethod}
\label{sec:method}
% 自\cite{snell2024scaling}提出test time scale up之后，推理模型不断涌现，一批优秀的推理模型例如ChatGPT-o1 (\cite{chatgpto1})、QwQ-32B (\cite{qwq32})以及DeepSeek-R1 (\cite{guo2025deepseek})等不断涌现，展现出测试时间扩展对模型提升的卓越贡献。基于此，我们希望引入推理方法来提升MLLMs对scientific cover image理解能力。
% Since test-time scale-up was introduced by~\cite{snell2024scaling}, a series of inference models have continuously emerged. Notable examples such as ChatGPT-o1~\citep{chatgpto1}, QwQ-32B~\citep{qwq32} and DeepSeek-R1~\citep{guo2025deepseek} have demonstrated the remarkable improvements that test-time scaling can contribute to model performance. Building on these advances, we aim to incorporate inference methods to enhance the ability of MLLMs in understanding scientific cover images. In this context, we propose our approach, \textbf{\ourMethod (\ourMethodShort)}.
% Since the introduction of test-time scale-up by~\cite{snell2024scaling}, a series of inference models—such as ChatGPT-o1~\citep{chatgpto1}, QwQ-32B~\citep{qwq32}, and DeepSeek-R1~\citep{guo2025deepseek}—have demonstrated their potential to significantly boost model performance. Building on these advances and recognizing that traditional in-context learning methods are less effective for scientific understanding tasks (\tabref{tab:icl}), we propose \textbf{\ourMethod (\ourMethodShort)}, a two-stage inference-time approach designed to enhance MLLMs’ comprehension of scientific cover images.

We first experiment with traditional in-context learning methods and find that they are ineffective for complex scientific understanding tasks (\tabref{tab:icl} and \tabref{tab:publisher}).
Motivated by the inference-time scaling~\cite{snell2024scaling} and a series of powerful inference models, such as ChatGPT-o1~\citep{chatgpto1}, QwQ-32B~\citep{qwq32}, and DeepSeek-R1~\citep{guo2025deepseek},
we propose \textbf{\ourMethod (\ourMethodShort)}, a two-stage inference-time approach specifically designed to enhance MLLMs’ comprehension of scientific cover images.

In the first stage, the full cover image and its classification question are fed into an MLLM to generate a detailed image description and a pseudo chain-of-thought~\citep{huang2025vision}, which guides reasoning without revealing the answer. Crucially, both DAD and MLLM approaches process all images simultaneously in a single batch rather than sequentially, ensuring that any performance variations stem solely from the introduction of the reasoning model, thereby maintaining evaluation fairness. In the second stage, these outputs are combined with the question text and passed to a dedicated reasoning model that produces a probability distribution over the options. By bridging visual perception with high-level reasoning, this structured process improves interpretability and equips MLLMs with the ability to reason about deeper connections between visual elements and scientific concepts.

\section{Experiments}

\begin{table}[t]
\centering
\renewcommand{\arraystretch}{1.3}

\adjustbox{max width=\linewidth}{
\large
    \begin{tabular}{ccccccccc}
    \toprule
    \multirow{3}{*}{\textbf{MLLMs}} & \multicolumn{4}{c}{\textbf{Image2Text Level}}                       & \multicolumn{4}{c}{\textbf{Text2Image Level}}                       \\
        & \multicolumn{2}{c}{Info Domain} & \multicolumn{2}{c}{Option Domain} & \multicolumn{2}{c}{Info Domain} & \multicolumn{2}{c}{Option Domain} \\
        & Acc.(\%)$\uparrow$       & ECE $\downarrow$               & Acc.(\%)$\uparrow$      & ECE $\downarrow$                & Acc.(\%)$\uparrow$       & ECE $\downarrow$               & Acc.(\%)$\uparrow$             & ECE $\downarrow$           \\ \hline
    Qwen2.5-VL-7B                   & 59.7      & 0.055        & 60.5       & \underline{ 0.063}         & 57.1      & 0.120              & 61.0            & 0.131          \\
    Qwen-VL-Max                        & 69.8      & 0.171              & 69.6       & 0.180               & 70.4      & 0.214              & 72.6            & 0.232          \\
    Step-1V-8k                      & 67.6      & 0.181              & 69.0       & 0.138               & 64.4      & 0.124              & 65.1            & 0.117          \\
    Step-1o-T-V          & 70.4      & 0.061              & 68.7       & 0.079               & 69.6      & 0.083              & 71.5            & 0.116          \\
    Ernie-4.5-8k                    & 57.5      & 0.231              & 61.4       & 0.158               & 55.8      & 0.095              & 58.3            & \underline{0.053}          \\
    Gemini-1.5-Pro                  & 72.7      & 0.108              & 70.4       & 0.085               & 71.8      & 0.237              & 72.8            & 0.172          \\
    GPT-4o                          & \underline{75.1}      & \textbf{0.053}     & \underline{73.5}       & \textbf{0.055}      & \underline{74.3}      & \textbf{0.038}     & \underline{74.3}            & 0.068          \\
    Step-3 & \textbf{77.3}               & 0.069              & \textbf{75.4}               & \underline{0.057}               & \textbf{78.0}               & \underline{0.065}               & \textbf{79.1}               & \textbf{0.048}               \\\bottomrule
    \end{tabular}
}
\caption{\textbf{Performance on \latestBench.} The table presents the experimental results of MLLMs on the understanding tasks of \latestBench, with \textbf{bold} indicating the best results and \underline{underlined} indicating the second-best results. Step-1o-T-V represents Step-1o-Turbo-Vision.}
% 表格中是对基线模型在\myBench的understanding tasks中的实验结果，其中加粗为最优结果，下划线为次优结果。
\label{tab:understanding-result}
\end{table}

We first introduce the baseline models and evaluation metrics (\secref{subsec:setting}), followed by the understanding evaluation of these state-of-the-art models and the performance gains achieved through our \ourMethod framework on \latestBench (\secref{subsec:understanding}). We next assess the effectiveness of both baseline models and our method on the text-free version of \latestBench (\secref{subsec:text-free}). Finally, we analyze the impact of different reasoning models integrated into our framework (\secref{subsec:reasoning-ablation}).
% 之后我们测试了基线模型和我们的方法在text-free version \latestBench上的效果(\secref{subsec:text-free})，最后我们分析了不同推理模型在我们方法上的表现。

% This section provides a detailed description of the experimental setup (\secref{subsec:setting}), including the evaluation performance of the state-of-the-art MLLMs and our \ourMethod framework optimization on our \latestBench (\secref{subsec:understanding}), as well as an analysis of various model error cases in the experiments (\secref{subsec:error}). Finally, we conduct ablation experiments on the dataset and the reasoning model in our proposed \ourMethod, and supplement the in-context learning prompting experiments for the models (\secref{subsec:ablation}).

\subsection{Settings}
\label{subsec:setting}

\paragraph{Models}

% We first surveyed current open-source and closed-source multimodal large language models (MLLMs). Across multiple capability evaluations, we observed that closed-source models significantly outperform their open-source counterparts~\citep{Guan_2024_CVPR,yue2024mmmu,chen2024we}. Consequently, for baseline models, we primarily selected mainstream commercial closed-source MLLMs, including Qwen-VL-Max~\citep{bai2023qwen}, Step-1V-8k, GPT-4o, Gemini-1.5-Pro~\citep{team2023gemini,team2024gemini}, and ernie-4.5-8k. For comparison, we also present the performance of recent state-of-the-art open-source model Qwen2.5-vl-7B~\citep{bai2025qwen2} and visual reasoning model Step-1o-turbo-vision on our benchmark. For the CoVR method, we employed QwQ-32B~\citep{qwq32} as its reasoning model to evaluate the benefits of test-time scaling.
We start by reviewing both open-source and closed-source multimodal large language models (MLLMs), finding that closed-source models consistently outperform their open-source counterparts across multiple benchmarks~\citep{Guan_2024_CVPR,yue2024mmmu,chen2024we,su2025essayjudge}. In light of the observed performance gap, our baselines primarily include leading commercial closed-source models: Qwen-VL-Max~\citep{qwen_max}, Step-1V-8k~\citep{step1v}, GPT-4o~\citep{hurst2024gpt}, Gemini-1.5-Pro~\citep {team2024gemini}, and Ernie-4.5-8k~\citep{ernie}. For comparison, we also report results of the open-source model Qwen2.5-vl-7B~\citep{bai2025qwen2} and the visual reasoning model Step-1o-Turbo-Vision~\citep{step1o}. To reflect the latest advances in multimodal models, we also evaluate Step-3, the strongest open-source multimodal reasoning model concurrent to our work. We leverage QwQ-32B~\citep{qwq32} as the reasoning model in \ourMethod (\ourMethodShort) to evaluate the effects of inference-time scaling.

\paragraph{Metrics}
In our experiments, we report 4 metrics of each MLLM on \latestBench to reflect the models’ classification performance and confidence performance on our benchmark:
\begin{itemize}
    \item  \textbf{Accuracy (Acc.)} measures the proportion of correct predictions, reflecting models' ability to identify correct scientific concepts.
    % $\text{Acc.} = \frac{1}{N}\sum_{i=1}^N \mathds{1}(\hat{y}i = y_i)$
    \item \textbf{Expected Calibration Error (ECE)}~\citep{guo2017calibration} evaluates prediction reliability, measuring whether models' confidence levels match their actual accuracy.
    % $\text{ECE} = \sum_{m=1}^M \frac{|B_m|}{n} |\text{acc}(B_m) - \text{conf}(B_m)|$~\citep{guo2017calibration}$
    \item \textbf{Negative Log-Likelihood (NLL)}~\citep{guo2017calibration} assesses prediction quality, penalizing both incorrect predictions and misaligned confidence, with higher penalties for high-confidence errors.
    % $\text{NLL} = -\frac{1}{N}\sum_{i=1}^N \log p(y_i|\mathbf{x}i)$~\citep{guo2017calibration}$
    \item \textbf{Root Mean Square Error (RMS)}~\citep{hendrycks2018deep} quantifies prediction error, measuring the magnitude of prediction deviations.
    % $\text{RMS}=\sqrt{\sum_{i=1}^b \frac{|B_i|}{n} \bigg( \frac{1}{|B_i|}\sum_{k\in B_i}\mathds{1}(y_k = \widehat{y}_k) - \frac{1}{|B_i|}\sum_{k\in B_i} c_k \bigg)^2}.$
\end{itemize}

\subsection{Understanding Evaluation}
\label{subsec:understanding}

\begin{figure}[t]
    \centering
    \includegraphics[width=1\linewidth]{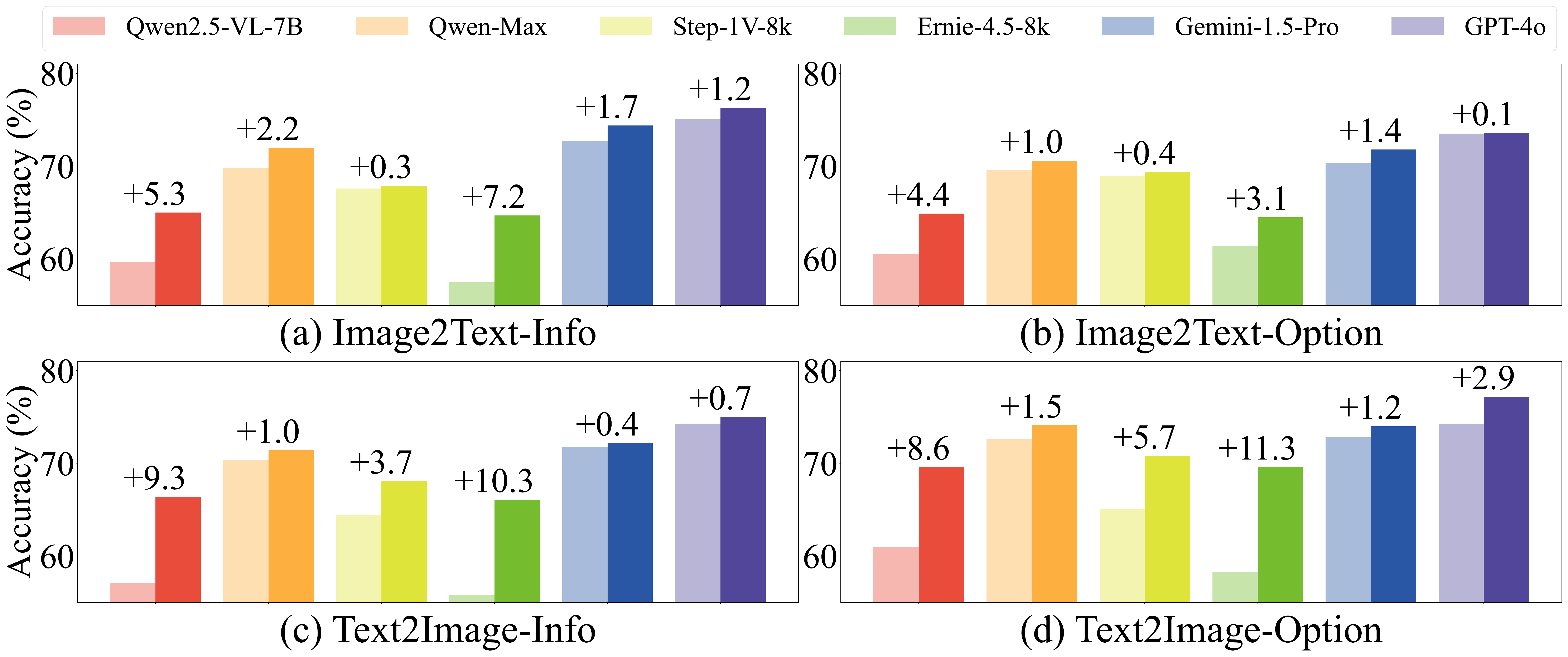}
    \caption{
    \textbf{Improvements of our \ourMethodShort across MLLMs.}
    Different colors represent different models, with the darker bars next to each model indicating the effect of our method.
    \ourMethodShort is generally effective in all four settings, shown in Subfigures (a)-(d).
    Best viewed in color.
    }
    \label{fig:ours}
\end{figure}
\Tabref{tab:understanding-result} and \figref{fig:ours} present classification results under four experimental settings, where models must identify visual scientific elements and extract relevant scientific concepts, aligning them with concepts in corresponding cover stories. We observe that these state-of-the-art MLLMs achieve accuracy rates between 50\% and 80\% on our \latestBench, highlighting the benchmark’s strong discriminative capacity in evaluating models’ comprehension of complex, cutting-edge scientific content. This suggests that existing models still face notable challenges in processing multimodal information and performing higher-level scientific reasoning. However, our proposed method \ourMethodShort, whose result is shown in \figref{fig:ours}, serving as a slight-weight inference-time approach, significantly improves the performance of these MLLMs, particularly achieving an 11.3\% accuracy gain for Ernie-4.5-8k in the option domain of the \ttoi task. Through effective cross-modal information fusion and optimization of reasoning mechanisms, our approach enhances each model's understanding and reasoning capabilities regarding scientific concepts, demonstrating robust performance.
% \Tabref{tab:understanding-result} and \figref{fig:ours} show classification results under four experimental settings. The task requires models to identify scientific elements in the cover image and extract relevant concepts from both image and story, establishing semantic connections across modalities. Current state-of-the-art MLLMs achieve 50–75\% accuracy on \latestBench, demonstrating the benchmark’s effectiveness in evaluating comprehension of complex scientific content. This indicates that models still struggle with multimodal understanding and high-level reasoning. In contrast, our inference-time method \ourMethodShort (results in \figref{fig:ours}) significantly improves performance by effectively fusing cross-modal information and enhancing reasoning, showing strong and consistent gains.

\subsection{Performance on Text-Free Cover Image}
\label{subsec:text-free}
\begin{figure}[htb]
    \centering
    \includegraphics[width=1\linewidth]{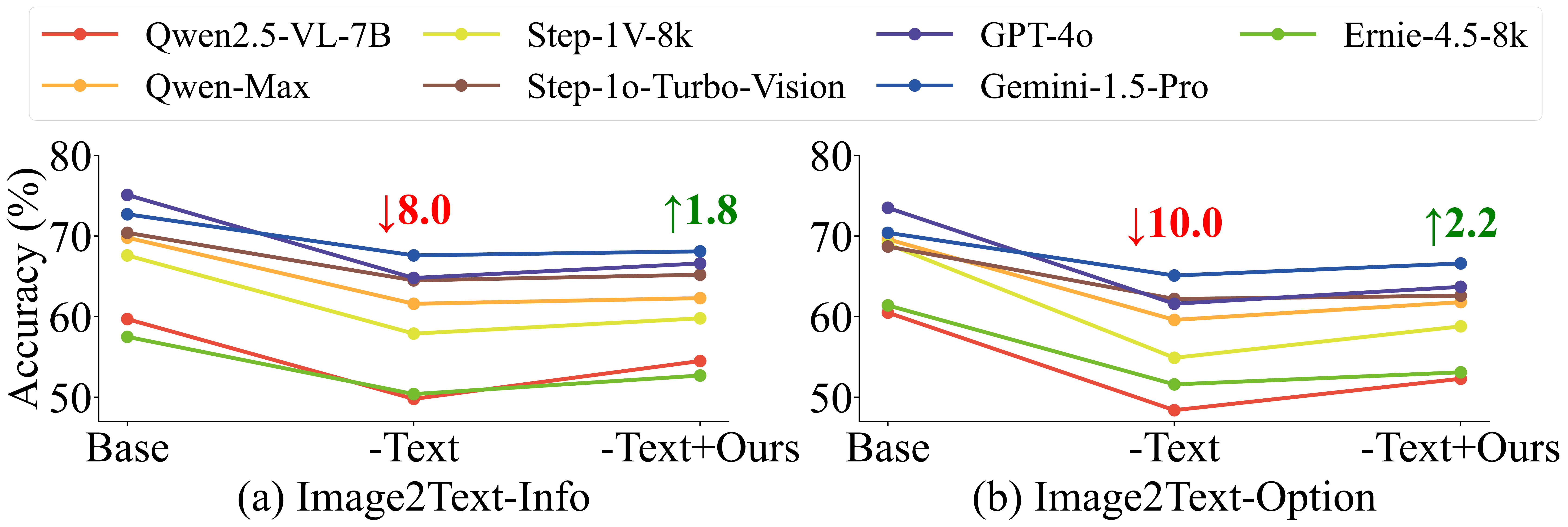}
    \caption{
    \textbf{Performance on covers with or without texts.}
    \say{-Text} indicates the removal of text from cover images using OCR on \latestBench .
    \say{+Ours} refers to our \ourMethodShort instead of simple MLLMs.
    Removing text leads to a significant performance drop, while our \ourMethodShort approach remains robust, even without relying on textual content.
    Best viewed in color.
    }
    \label{fig:text-free}
\end{figure}
Since MLLMs may rely on rich textual elements, such as journal titles, author names, and detailed annotations, rather than visual content to comprehend scientific cover images, we design a controlled experiment to evaluate their ability to understand scientific concepts based solely on visual information. To this end, we systematically apply OCR~\citep{EasyOCR} to the original images to identify and remove the main textual regions, subsequently masking them with white blocks to eliminate potential textual cues while preserving the overall visual structure. We set the confidence threshold for this process to 0.25. This value is determined by experimenting with thresholds ranging from 0.1 to 0.9 on 10 randomly sampled cover images from four different publishers. The final value is chosen by considering both the effectiveness of text removal and the preservation of key visual-semantic elements in the images. We then assess several representative models under this text-free setting and conduct an ablation experiment where text is removed from images but provided in the prompt to determine whether performance degradation is due to image distortion.

As shown in \figref{fig:text-free}, removing text causes a notable drop in accuracy, revealing that current models struggle to reason about complex scientific concepts from visual input alone, while \tabref{tab:ocr} indicates that the performance degradation is not due to image distortion. The case study in \figref{fig:case_study} demonstrates that even the most advanced model released concurrent to our work, Gemini 2.5 Pro~\citep{gemini2.5pro}, struggles to reason from perceived visual elements to complex scientific concepts. In contrast, our \ourMethodShort method significantly improves performance by leveraging test-time scaling and relying solely on the initial visual grounding step. These results indicate that while MLLMs excel at perceiving elements in cover images, their visual-only scientific understanding remains limited, yet can be effectively enhanced through structured reasoning mechanisms.

\subsection{Reasoning Module Comparison in \ourMethodShort}
\label{subsec:reasoning-ablation}
% To further validate the performance of different reasoning models within the \ourMethod framework, we selected several representative reasoning models, including Deepseek-R1 and ChatGPT-o3-mini, as alternatives to the QwQ-32B reasoning model used in the original experiments. 
% Experimental results in \tabref{tab:reasoning-ablation} reveal significant performance variations across different language models within the \ourMethod framework. The introduction of more advanced reasoning models, such as Deepseek-R1, significantly improved accuracy even for the most capable MLLMs. The integration of reasoning models through the \ourMethodShort method demonstrated high accuracy and robustness in image-to-text reasoning tasks. These findings further validate the importance of inference-time scaling for multimodal understanding tasks while providing guidance for future MLLM optimization.

To further assess the impact of different reasoning models within our \ourMethod framework, we replaced the original QwQ-32B reasoning model with several representative alternatives, including Deepseek-R1~\citep{guo2025deepseek} and ChatGPT-o3-mini-high~\citep{chatgpto3}. As shown in \tabref{tab:reasoning-ablation}, the results reveal notable performance differences across reasoning models. Incorporating more advanced models, such as Deepseek-R1, leads to significant accuracy gains, even for already strong MLLMs. The integration of reasoning models via the \ourMethodShort method consistently yields high accuracy and robustness in image-to-text tasks. These findings underscore the importance of inference-time scaling in multimodal understanding and offer valuable insights for optimizing future MLLMs.

% \subsection{Make \latestBench Alive}
% 在前文中我们论述了通过持续的数据更新使benchmark保持活力，本节将阐述如何通过先进的embedding模型提升benchmark的挑战性。随着多模态embedding模型的不断发展，这些模型在静态benchmark上展现出持续提升的性能。为深入研究这一现象，我们选取了与我们的工作同时提出的三个embedding model：SigLip2、Qwen-multimodal-embedding-v1和doubao-embedding-vision-241215。
% 我们利用这些先进的embedding模型重新构建了选择题，同时保持与MAC-2025相同的所给信息集合，确保新的benchmark仅在干扰项的选择上有所差异。这种方法确保了评估的一致性，同时允许我们精确衡量embedding模型选择对benchmark难度的影响。为评估这一方法的有效性，我们选取了三个具有代表性的baseline模型，在image2text任务的given info domain进行了实验。实验结果显示，采用更为先进的embedding模型构建的benchmark能够显著降低MLLMs的表现：所有测试模型的准确率平均下降了1\%-2\%，其中最显著的性能下降达到了x.xx\%。同时，每个模型的ECE和RMS表现均有所下降，这表明新的benchmark不仅降低了模型的准确率，还影响了其置信度预测的可靠性。
% 这一结果具有重要意义：首先，它证实了我们的观点，即更先进的embedding模型能够捕捉更细微的语义差异，从而生成更具挑战性的干扰项；其次，这种性能下降表明，即使是当前最先进的MLLMs在处理细微的语义差异时仍存在局限性；最后，这种方法为构建动态适应性benchmark提供了新的思路，使得评估体系能够与模型能力的发展保持同步。
% 这些发现不仅验证了我们方法的有效性，也为未来benchmark的设计和优化提供了重要参考。随着embedding模型的持续进步，我们预期这种方法将能够持续保持其挑战性，为MLLMs的评估提供更加可靠和有意义的标准。

\section{\BenchFullShort's \emph{Live} Attribute}

To rigorously demonstrate the \emph{live} attribute of our \BenchFull, we conduct experiments on two perspectives, \emph{live} data and \emph{live} data curation.
In \secref{subsec:live_data}, we demonstrate that newer data, represented by the MAC-2025 subset, poses significantly greater challenges than earlier subsets, underscoring the increased complexity introduced by cutting-edge scientific knowledge.
In \secref{subsec:live_curation}, we further amplify the benchmark’s difficulty by curating distractors using contemporary embedding models.
Our controlled experiments show that stronger embeddings yield more confusing distractors, resulting in substantial accuracy drops even for state-of-the-art MLLMs.
% Together, these findings illustrate how live data and live curation synergize to maintain a challenging, adaptive, and future-proof benchmark for scientific understanding.

\subsection{\emph{Live} Data}
\label{subsec:live_data}
\begin{figure}[t]
    \begin{minipage}[b]{0.62\textwidth}  % 图片占40%的宽度
        \centering
        \includegraphics[width=\linewidth]{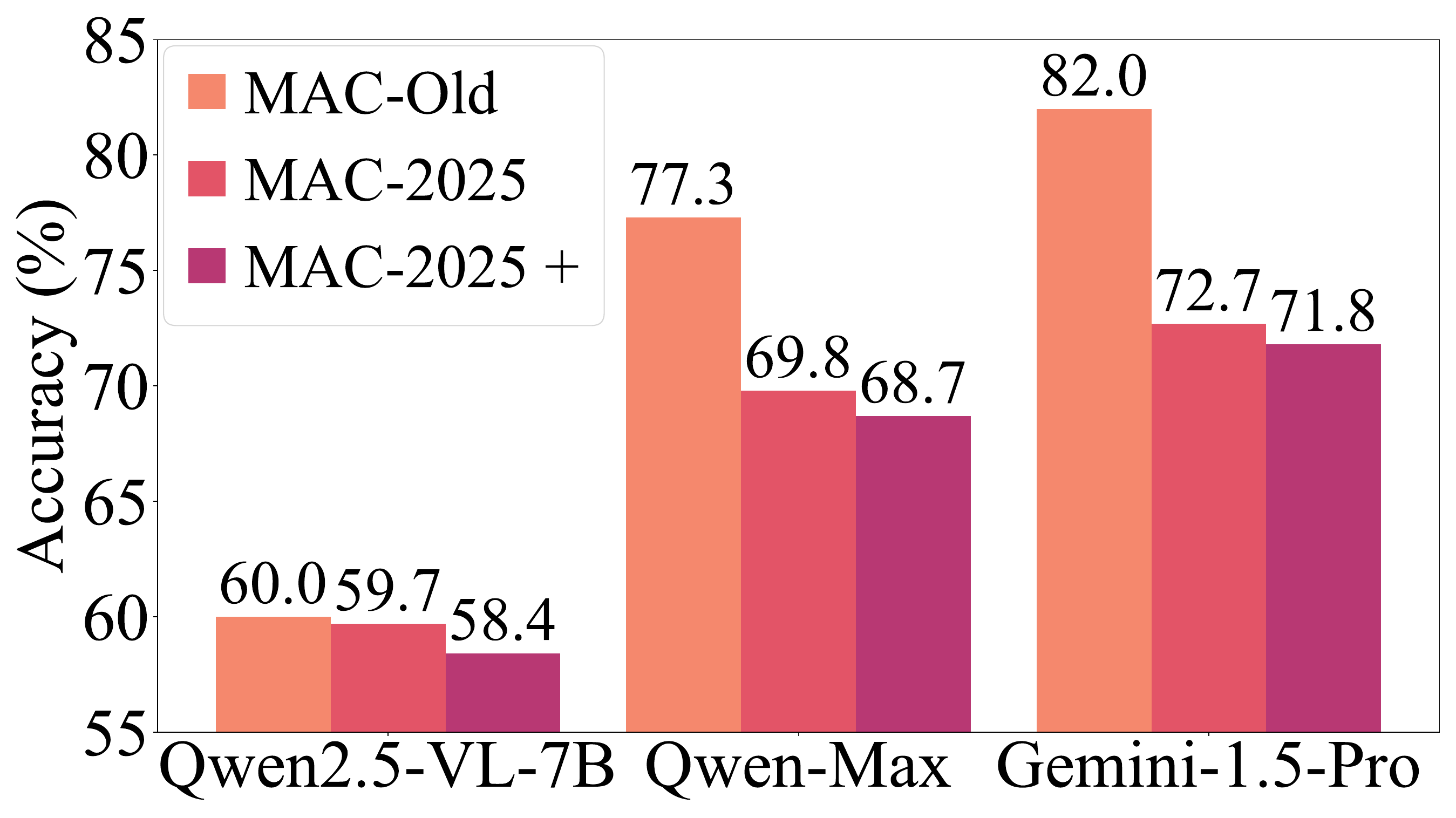}
    \end{minipage}%
    \hfill  % 插入空白使文字与图片分开
    \begin{minipage}[b]{0.33\textwidth}  % 图注占55%的宽度
        \caption{\textbf{Benchmark with distinct data and models for curation across different periods.} 
        MAC-Old and \latestBench use the oldest and latest journal issues, respectively.
        \say{\latestBench +} is curated by the latest embedding models.
        Our \BenchFullShort is getting more challenging, benefited from \emph{live} data and \emph{live} data curation.
        }
        \label{fig:live}
    \end{minipage}
\end{figure}

To demonstrate that recent data poses greater challenges to current MLLMs, we evaluate model performance on over 2,000 samples drawn from the earliest available year across all journals in \BenchFullShort, shown as the \BenchFullShort-Old in \figref{fig:live}. Comparative analysis shows that models perform significantly worse on \latestBench than on earlier subsets \BenchFullShort-Old, indicating that incorporating newer, cutting-edge scientific discoveries substantially increases task difficulty (\tabref{tab:oldest}). This temporal performance degradation can be attributed to the fact that recent scientific advances often introduce novel concepts, methodologies, and terminology that fall outside the training distributions of current models, creating inherent challenges for comprehension and reasoning. This notable temporal performance gap further supports our \emph{live} design philosophy: by continuously integrating the latest scientific advances, the benchmark preserves both its relevance and difficulty. Our \emph{live} \latestBench leverages up-to-date frontier scientific knowledge to evaluate models' scientific understanding, ensuring consistent and fine-grained evaluation of their capabilities.
% \begin{figure}[htb]
%     \centering
%     \includegraphics[width=0.5\linewidth]{Figure/live.pdf}
%     \caption{\textbf{Performance of baseline models across datasets from different periods and benchmarks constructed using different distractor selection models.} For each model, the hollow bars represent NLL performance, while the solid bars indicate accuracy.}
%     \label{fig:live}
% \end{figure}

\subsection{\emph{Live} Data Curation}
\label{subsec:live_curation}
% Building upon our previous discussion of maintaining benchmark vitality through continuous data updates, this section demonstrates how evolving embedding models can enhance the benchmark's challenging nature. As multimodal embedding models continue to evolve, they consistently demonstrate improved performance on static benchmarks. For an in-depth examination of this phenomenon, we selected three contemporary embedding models introduced concurrent with our work: SigLip2, Qwen-multimodal-embedding-v1, and doubao-embedding-vision-241215.

% We leveraged these advanced embedding models to reconstruct classification questions in \latestBench while maintaining the same given information set, ensuring that the new benchmark differs only in its distractor options. This approach ensures evaluation consistency while enabling precise measurement of how \emph{live} data curation impacts benchmark difficulty. We selected three representative baseline models and conducted experiments in the given info domain of the image2text task. The results in \figref{fig:live} demonstrate that benchmarks constructed using more sophisticated embedding models significantly increases model confusion and makes its predictions less reliable.
To reveal how evolving embedding models enhance benchmark's challenging nature, we rebuilt distractors of the classification questions in \latestBench using three contemporary embedding models introduced concurrently with our work—SigLip2~\citep{tschannen2025siglip}, Qwen-Multimodal-Embedding-V1~\citep{qwen_embedding}, and doubao-embedding-vision-241215~\citep{doubao_embedding}—while keeping the original information set fixed. This controlled experimental setup allows us to isolate the impact of updated distractors on evaluation difficulty while maintaining consistency in the underlying knowledge being tested. The selection of these three embedding models represents newer and more powerful iterations that demonstrate significant improvements over previous generations, ensuring that our analysis captures how enhanced embedding capabilities influence distractor quality. Furthermore, by maintaining identical correct answers and question structures across all experimental conditions, we establish a rigorous framework for attributing performance variations solely to the sophistication of the distractor generation process. Experiments on the \itot task with three baseline models (see results in \figref{fig:live} and details in \tabref{tab:mac2026}) show that stronger embedding models produce more confusing distractors, leading to notable accuracy drops and less reliable predictions.

These findings have several important implications: First, they validate our hypothesis that more advanced embedding models can generate more challenging distractor items; Second, the performance degradation indicates that even state-of-the-art MLLMs still have limitations in processing subtle semantic differences; Finally, this approach provides new insights into constructing dynamically adaptive benchmarks that evolve alongside model capabilities.
Our \emph{live} \latestBench utilizes evolving multimodal embedding models to construct distractors matched to the current capabilities of MLLMs, ensuring sustained challenge and reliability while offering valuable insights for future benchmark design and optimization.
\section{Conclusion}
In this work, we introduce \BenchFull, a live benchmark that dynamically evolves alongside scientific advances and model progress to evaluate MLLMs' scientific understanding capabilities. Leveraging over 25,000 carefully curated image-text pairs from leading scientific journals and concentrating on our latest snapshot \latestBench, we reveal a critical finding: while current MLLMs demonstrate strong visual perception abilities, they exhibit significant limitations in cross-modal scientific reasoning. To address this fundamental gap, we propose \ourMethod, a lightweight inference-time approach that effectively bridges visual cues with language-based reasoning processes, achieving substantial performance improvements across multiple scientific domains. Through rigorous temporal generalization analysis and systematic evolving benchmark construction, we further demonstrate the necessity and feasibility of live evaluation frameworks, highlighting the critical importance of continuously adaptive benchmarks for accurately monitoring MLLMs' scientific comprehension capabilities as they advance over time.

Looking forward, we envision several promising directions for advancing scientific MLLM evaluation. A key priority lies in developing more fine-grained evaluation metrics that can systematically distinguish between perception-understanding and cognition-reasoning abilities. Specifically, we plan to design granular assessment frameworks that isolate visual perception tasks such as identifying scientific items and extracting textual information from higher-order cognitive reasoning tasks including causal inference, hypothesis formation, and scientific concept understanding. This differentiation will enable more precise diagnosis of model capabilities and targeted improvements in specific areas. Additionally, we aim to expand our evaluation paradigm beyond discriminative tasks to include generation-based assessments, where models must produce coherent scientific explanations and demonstrate reasoning processes. We believe that such comprehensive and evolving benchmarks will not only provide deeper insights into current model limitations but also serve as crucial catalysts for the development of more sophisticated MLLMs, ultimately driving the field toward systems capable of genuine scientific discovery and reasoning.

% Moving forward, we plan to extend MAC beyond accuracy metrics to incorporate fine-grained evaluation that systematically distinguishes between perception-understanding and cognition-reasoning abilities of MLLMs. Such granular assessment would provide deeper insights into specific model strengths and weaknesses.
% Additionally, exploring other domains with evolutionary characteristics similar to science, such as technology and current events analysis, can offer exciting opportunities. Developing live benchmarks in these rapidly evolving areas would create a comprehensive framework for continuously challenging MLLM capabilities across multiple knowledge-intensive domains.

\section*{Acknowledgement}
This research is supported by the Key R\&D Program of Shandong Province, China (2023CXGC010214, 2024CXGC010213). We express our gratitude to the funding agency for their support. We thank all the anonymous reviewers for their valuable suggestions.

% \bibliography{colm2025_conference}
% \bibliographystyle{colm2025_conference}

\bibliography{ref}
\bibliographystyle{colm2025_conference}

\appendix
\newpage
\section{Open Access License of the Journals}
% 我们在构建 \Dataset 时使用的所有封面图像及其对应的封面故事，均来自顶级科学期刊的官方网站页面，这些期刊包括 Nature 、Science/AAAS、Cell Press 以及美国化学会（ACS）出版的期刊。我们严格确保所有纳入基准测试的数据内容均符合各出版方所声明的开放获取或公共使用政策。所有封面图像与封面故事的使用仅限于非商业的学术研究。我们不对原始文章进行再分发，也不对原始内容进行修改或宣称其所有权。在适用场景中，我们保留了必要的署名信息，整个数据集的构建过程也遵循“合理使用”原则，旨在支持科学评测与研究可复现性。
All cover images and their corresponding cover stories used in the construction of \BenchFull~(\BenchFullShort) are sourced from the official websites of leading scientific journals which published by \href{https://www.nature.com/nature-portfolio/editorial-policies/self-archiving-and-license-to-publish#creative-commons-licences}{Nature Portfolio}, \href{https://www.science.org/content/page/science-licenses-journal-article-reuse}{Science/AAAS}, \href{https://www.cell.com/cell/authors}{Cell Press} and \href{https://solutions.acs.org/licensing-overview/}{American Chemical Society}. We strictly ensure that all content included in the \BenchFullShort complies with the open access or public use policies stated by each publisher.

% The use of cover images and cover stories is strictly limited to non-commercial academic research. We do not redistribute full journal articles, nor do we modify or claim ownership of any original content. Proper attribution is preserved where applicable, and the entire dataset construction process adheres to fair use principles, aiming to support scientific evaluation and research reproducibility.

The use of cover images and their associated cover stories is strictly limited to non-commercial academic research purposes. We do not redistribute full journal articles, nor do we modify, alter, or claim ownership of any original content. Proper attribution and source referencing are preserved wherever applicable, and the entire dataset construction process adheres to established fair use principles, with the goal of supporting rigorous scientific evaluation, reproducibility, and transparency in academic research.

\section{Self-Validation Experiment Result}
\label{sec:self-validation}
\begin{table}[htb]
\centering
\renewcommand{\arraystretch}{1.2}
\adjustbox{max width=\linewidth}{
    \begin{tabular}{ccccccc}
    \toprule
    \textbf{Modal}              & \textbf{Embedding Model} & \textbf{Average$\downarrow$} & \textbf{Median$\uparrow$} & \textbf{Acc.(\%)$\uparrow$} \\ \midrule
    \multirow{4}{*}{Image2Text} & CLIP-L                   & 0.255            & 0.150           & 16.2         \\
                                & ViT-H                    & 0.227            & 0.105           & \textbf{23.9}         \\
                                & ViT-G                    & 0.287            & 0.193           & \underline{16.7}         \\
                                & average                  & 0.237            & 0.128           & 21.6         \\ \hline
    \multirow{4}{*}{Text2Image} & CLIP-L                   & 0.203            & 0.044           & 32.0         \\
                                & ViT-H                    & 0.205            & 0.037           & \underline{35.8}         \\
                                & ViT-G                    & 0.238            & 0.076           & 29.2         \\
                                & average                  & 0.194            & 0.034           & \textbf{38.7}         \\\bottomrule
    \end{tabular}
}
\caption{\textbf{Analysis of multimodal Embedding Models' Performance in Scientific Understanding Tasks and Distractor Selection. }We assessed the models through two metrics: self-ranking (similarity rankings of paired stories or images among all journal samples) and accuracy rates in scientific comprehension multiple-choice tasks. }
% 表格中是对八个闭源模型和两个开源模型在\myBench的understanding task中的实验结果，其中上标星号(*)为开源MLLM模型。
\label{tab:embedding-self-proof}
\end{table}
% \Tabref{tab:embedding-self-proof}中是对我们选取\latestBench时使用的embedding model的self-validation分析实验结果。其中：Average为各个embedding model在所有期刊中自相似度排名（即image的embedding与其对应pair的story的embedding的相似度在当前期刊中所有的issue的相似度排名比例），Median为自相似度的中位数排名，Acc.为embedding model回答已经筛选完成的\latestBench问题的正确率。Embedding model中的average为每个issue在三个embedding model的排名取平均后重新排序得到的相似度排名，综合了三个embedding model的特点。
\Tabref{tab:embedding-self-proof} presents the self-validation results of the embedding models used in selecting distractors for \latestBench. Specifically, ``Average" indicates the average relative rank of self-similarity for each embedding model across all journal issues—that is, the rank of similarity between the embedding of an image and its corresponding story, relative to all other candidate stories from the same journal. Median represents the median of these self-similarity rankings. Acc. refers to the model’s accuracy in answering the finalized \latestBench classification questions using similarity alone. The “average” row under each modality corresponds to a combined ranking, obtained by averaging the ranks produced by the three embedding models for each issue and re-ranking accordingly, thus reflecting a composite similarity measure that integrates their respective strengths.

% 从\tabref{tab:embedding-self-proof}我们发现，
We evaluate the capability of three embedding models (CLIP-L, ViT-H, and ViT-G) on our bidirectional matching tasks using similarity-based retrieval, as reported in \tabref{tab:embedding-self-proof}. Results reveal a consistent performance gap between the Image2Text and Text2Image tasks. Specifically, the average accuracy across models for Image2Text is 21.6\%, with an average rank of 0.237, whereas Text2Image achieves a substantially higher average accuracy of 38.7\% and a lower average rank of 0.194. This indicates that retrieving matching images from text is generally easier than retrieving text from images, likely due to the higher semantic complexity embedded in scientific visual content. Among the models, ViT-H consistently outperforms others across both modalities, achieving the highest accuracy in Image2Text (23.9\%) and Text2Image (35.8\%), along with competitive average ranks (0.227 and 0.205, respectively). In contrast, ViT-G performs the worst, with lower accuracy and higher rank values in both tasks, suggesting limited effectiveness in capturing cross-modal semantic alignment. Despite the relatively low accuracy scores—particularly in the Image2Text setting—all models rank the correct answer within the top 25\% on average, indicating that the selected distractors are semantically close and thus effective for challenging MLLMs.
\section{Human Evaluation}
% human eval的数据放到这里
To evaluate human performance on our benchmark, we selected 20 journal covers from diverse publishers in the \latestBench dataset and designed two evaluation configurations: (1) random distractors and (2) embedding-based distractors consistent with the image2text info domain methodology employed in \latestBench.

We recruited five independent evaluators, each with at least a bachelor's degree, to complete the assessment tasks. The results demonstrate notable performance differences: human evaluators achieved 80\% accuracy with random distractors, while accuracy decreased to 76\% with embedding-based distractors. This performance degradation provides empirical evidence for the effectiveness of our distractor design strategy. The significant accuracy reduction when using semantically-informed distractors validates our approach and confirms that the embedding-based methodology successfully increases task difficulty.
\section{\BenchFullShort-Old Benchmark Evaluation}
\begin{table}[htb]
\centering
\renewcommand{\arraystretch}{1.5}
\adjustbox{max width=\linewidth}{\large
\begin{tabular}{ccccccccc}
\toprule
\multirow{3}{*}{\textbf{MLLMs}} & \multicolumn{4}{c}{\textbf{Image2Text Level}}                       & \multicolumn{4}{c}{\textbf{Text2Image Level}}                       \\
                                & \multicolumn{2}{c}{info domain} & \multicolumn{2}{c}{option domain} & \multicolumn{2}{c}{info domain} & \multicolumn{2}{c}{option domain} \\
                                & Acc.(\%)$\uparrow$           & ECE$\downarrow$            & Acc.(\%)$\uparrow$            & ECE$\downarrow$             & Acc.(\%)$\uparrow$           & ECE$\downarrow$            & Acc.(\%)$\uparrow$            & ECE$\downarrow$             \\ \midrule
Qwen2.5-VL-7B                   & 60.0          & \textbf{0.042}          & 62.1           & \underline{0.068}           & 59.4          & \underline{0.126}          & 61.7           & \underline{0.147}           \\
Qwen-VL-Max                        & 77.3          & 0.196          & \underline{75.0}           & 0.196           & \underline{74.5}          & 0.244          & 75.5           & 0.273           \\
Gemini-1.5-Pro                  & \textbf{82.0}          & 0.149          & \textbf{80.6}           & 0.144           & 72.0          & 0.141          & \underline{76.0}           & 0.202           \\
GPT-4o                          & \underline{77.6}          & \underline{0.051}          & 74.8           & \textbf{0.058}           & \textbf{76.2}          & \textbf{0.071}          & \textbf{79.3}           & \textbf{0.083}  \\\bottomrule
\end{tabular}
}
\caption{\textbf{MLLMs' Performance on Benchmarks Generated Using MAC-Old Dataset. }On \BenchFullShort, four representative MLLMs achieve significantly higher accuracy compared to their performance on \latestBench, along with notably better confidence calibration.}
\label{tab:oldest}
\end{table}
% 在“Oldest”数据集上（即时间距离当前最久远的期刊内容），各大模型在四个任务分支（image2text/info、image2text/option、text2image/info、text2image/option）上的准确率整体优于“Latest”数据集，且置信度校准指标ECE也表现得更为稳定。例如，在image2text/info domain任务中，GPT-4o在“Oldest”上达到0.751的准确率和极低的0.053 ECE，而在“Latest”数据集上，同一任务准确率下降至0.643，ECE上升至0.094，表现明显下滑。

% 这一趋势在其他模型上也普遍存在，如Gemini-1.5-Pro在“Oldest”上的text2image/info任务准确率为0.718，而在“Latest”中仅为0.659；对应ECE也从0.237下降到0.212，显示了在较新的科学内容上模型面临更高的混淆性和理解难度。此外，Qwen2.5-VL-7B在text2image/info任务中从“Oldest”的0.571下降至“Latest”的0.476，差距显著，进一步说明基础模型对新兴知识的适应性仍有限。“Latest”数据集中所涵盖的科学发现具有更强的前沿性和表达复杂度，对MLLMs在科学语义理解和跨模态推理能力上提出了更高要求。这一现象验证了我们提出的\emph{live} benchmark设计理念，即通过持续纳入最新科学发现来确保基准测试始终处于人类知识的前沿，并保持对强大模型的有效挑战性。
All evaluated models achieve consistently higher accuracy on the \BenchFullShort-Old, which consists of journal issues furthest from the present, compared to their performance on the more recent \latestBench dataset. Moreover, the Expected Calibration Error (ECE) is generally lower and more stable. For instance, in the info domain of image2text, GPT-4o achieves an accuracy of 0.751 and an impressively low ECE of 0.053 on the \BenchFullShort-Old data. However, on the \latestBench dataset, its accuracy drops to 0.643 while the ECE rises to 0.094, reflecting a notable decline in both performance and calibration.
This trend is broadly consistent across other models. Gemini-1.5-Pro achieves 71.8\% accuracy in the text2image info task on \BenchFullShort-Old, but only 65.9\% on Latest; its ECE also increases from 0.212 to 0.237, indicating higher confusion and difficulty in interpreting more recent scientific content. 

The scientific discoveries presented in the latest dataset are more cutting-edge and linguistically complex, thereby posing greater and more nuanced challenges for MLLMs in both scientific semantic comprehension and cross-modal reasoning. These findings support our proposed design principle of a \emph{live} benchmark: by continually incorporating the latest scientific advances, the benchmark remains aligned with the frontier of human knowledge while sustaining its evaluative strength against powerful models.

\section{In-Context Learning}
\begin{table}[htb]
\centering
\renewcommand{\arraystretch}{1.5}
\begin{tabular}{ccccc}
\toprule
\multirow{2}{*}{\textbf{MLLMs}} & \multicolumn{4}{c}{\textbf{Image2Text Level}} \\
                       & Acc.(\%)$\uparrow$    & ECE$\downarrow$     & NLL$\downarrow$     & RMS$\downarrow$    \\ \midrule
Qwen2.5-VL-7B          & \underline{59.7}   & 0.055   & \textbf{1.856}   & \textbf{0.095}  \\
+ CoT                  & 56.6   & 0.184   & \underline{2.071}   & 0.223  \\
+ One-Shot             & 43.2   & \textbf{0.123}   & 4.136   & \underline{0.155}  \\
+ Self-Consistency(5)  & 59.2   & 0.214   & 2.897   & 0.285  \\
+ ours                 & \textbf{65.0}   & \underline{0.127}   & 2.729   & 0.162  \\\hline
Qwen-VL-Max            & 69.8   & 0.171   & 3.054   & 0.210  \\
+ CoT                  & \underline{69.8}   & \textbf{0.070}   & \textbf{1.517}   & \textbf{0.115}  \\
+ One-Shot             & 63.1   & 0.231   & 5.378   & 0.315  \\
+ Self-Consistency(5)  & 69.8   & 0.319   & \underline{2.143}   & 0.373  \\
+ ours                 & \textbf{72.0}   & \underline{0.082}   & 2.719   & \underline{0.117}  \\\bottomrule
\end{tabular}
\caption{\textbf{MLLMs' Performance on Benchmarks Generated Using State-of-the-Art Embedding Models. }We reconstructed new distractor items for samples identical to those in MAC-2025 from the MAC dataset using three state-of-the-art embedding models, and evaluated baseline models' performance on the image2text task.}
\label{tab:icl}
\end{table}
% 我们同时还在两个代表性模型 Qwen2.5-VL-7B 和 Qwen-Max 上尝试了通过 Chain-of-Thought (CoT)、few-shot 和 self-consistency 等 In-Context Learning 方法来提升 MLLMs 的性能。我们参考~\cite{wei2022chain} 和~\cite{wang2022self} 设置了 CoT 和 self-consistency 实验，并在 \tabref{tab:icl} 中报告了各方法在 Image2Text 任务上的表现。

% 从实验结果可见，CoT 方法对于 Qwen-vl-max 有一定的帮助，其 NLL 显著下降至 1.517，RMS 降至 0.115，说明其对模型信心判断具备一定稳定性；但在 Qwen2.5-VL-7B 上，CoT 不仅未带来准确率提升，反而出现 ECE 上升（由 0.055 升至 0.184）及 NLL 增大（由 1.856 升至 2.071），说明其在弱模型上可能引入更多不确定性。

% 在本任务上Few-shot 表现则相对不稳定：尽管在部分场景下能略微改善 ECE，但在两个模型上的准确率均明显下降，且 NLL 和 RMS 明显增大，尤其在 Qwen-vl-max 上 NLL 飙升至 5.378，显示出在复杂科学场景下 few-shot 泛化能力有限。5 runs的Self-consistency实验在 Qwen2.5-VL-7B 上略优于 few-shot，准确率维持在 0.592，但 ECE 与 RMS 明显高于原始模型。相似地，在 Qwen-vl-max 上 self-consistency 并未带来准确率提升，且在校准性上甚至劣于 few-shot（ECE 提升至 0.319）。

% 相比之下，我们提出的 \ourMethod 方法在两个模型上均取得了显著提升。Qwen2.5-VL-7B 准确率从原始的 0.597 提升至 0.650，Qwen-vl-max 从 0.698 提升至 0.720，且 ECE 和 RMS 均保持在较低水平，验证了我们提出的结构化推理机制在复杂视觉科学任务中的有效性和稳健性，表征了在MLLM中引入reasoning过程以带给MLLM intellect的重要潜力。
We further experimented with In-Context Learning strategies on two representative models, Qwen2.5-VL-7B and Qwen-Max, incorporating Chain-of-Thought (CoT), one-shot prompting, and self-consistency to enhance MLLM performance. Following the setups from~\cite{wei2022chain} and~\cite{wang2022self}, we implemented CoT and 5-sample self-consistency baselines, with results summarized in \tabref{tab:icl} for the Image2Text task.

% 补充一个不同domain下one-shot的实验表格

The results show that CoT offers some benefit for Qwen-Max, significantly reducing NLL to 1.517 and RMS to 0.115, indicating improved confidence calibration. However, for Qwen2.5-VL-7B, CoT not only fails to improve accuracy but also increases ECE (from 0.055 to 0.184) and NLL (from 1.856 to 2.071), suggesting that CoT may introduce instability in weaker models. One-shot prompting yielded unstable results for this task. While it slightly improved ECE in some cases, it consistently led to substantial drops in accuracy for both models and dramatic increases in NLL and RMS. For example, Qwen-Max’s NLL rose sharply to 5.378, indicating the limited generalization capacity of one-shot learning in complex scientific contexts. Self-consistency showed marginal improvement over few-shot on Qwen2.5-VL-7B, with accuracy reaching 0.592, though at the cost of higher ECE and RMS compared to the original model. On Qwen-Max, self-consistency failed to improve accuracy and even degraded calibration, with ECE increasing to 0.319.

In contrast, our proposed \ourMethod consistently delivered notable and robust gains. Qwen2.5-VL-7B’s accuracy improved from 0.597 to 0.650, and Qwen-Max from 0.698 to 0.720, while maintaining low ECE and RMS throughout. These results affirm the robustness and effectiveness of our structured reasoning mechanism in tracking complex scientific visual tasks and underscore the importance of incorporating explicit reasoning processes to endow MLLMs with deeper scientific understanding.

\begin{table}[htb]
\centering
\renewcommand{\arraystretch}{1.5}
\adjustbox{max width=\linewidth}{
\begin{tabular}{cccc}
\toprule
\textbf{MLLM}                  & \textbf{One-Shot} & \textbf{Acc.(\%)$\uparrow$} & \textbf{ECE} \\ \midrule
\multirow{4}{*}{Qwen2.5-VL-7B} & Cell              & \textbf{43.24}        & \underline{0.2313}       \\
                               & Science           & \underline{42.98}        & \textbf{0.2190}       \\
                               & Nature            & 40.62        & 0.2899       \\
                               & ACS               & 39.27        & 0.2960       \\ \bottomrule
\end{tabular}
}
\caption{\textbf{Model Performance on the Image2Text Task Using One-Shot Examples from Different Publishers.} We evaluated Qwen2.5-VL-7B on the \latestBench benchmark using one-shot examples from various publishers, and it demonstrated similar performance..}
\label{tab:publisher}
\end{table}
\section{Distractor Preference}
\label{subsec:error}

% 如\secref{subsec:benchmark}所述，对于\latestBench中选择题干扰项的选择，我们引入了多种不同训练数据集的embedding model来计算相似度。因此，为了分析各个MLLMs所出现的错误选项，我们检查了这些错误选项来源于哪些embedding model，并进一步分析了这些模型是否展现出对不同embedding model的偏好性。
As described in \secref{subsec:benchmark}, for the selection of distractors in the multiple-choice questions of \latestBench, we introduced various embedding models trained on different datasets to calculate similarity. Therefore, to analyze the incorrect options presented by MLLMs, we examined which embedding models these incorrect options originated from and further analyzed whether these models exhibited preferences for specific embedding models. By comparing the distribution of incorrect options, we aim to reveal the differences in how various embedding models select distractors, especially whether certain embedding models tend to frequently choose specific types of distractors when handling the same question.
% 这种偏好可能会影响模型的决策质量和准确性，尤其是在多模态任务中，其中图像和文本信息需要通过不同的方式进行对齐与推理。
% \Tabref{tab:error-ranking}展示了多个先进多模态大模型（如 GPT-4o、Gemini-1.5-Pro、Qwen-Max、Step 系列等）在四个任务维度上，不同嵌入模型(见\secref生成的干扰项下的错误率表现。从图中可以看出，不同模型在四个任务上的整体表现存在明显差异，尤其是在中心“Average”维度，模型之间的错误率差异清晰，说明各模型在面对不同模态匹配任务时的易混淆程度存在系统性差异。其中，Qwen2.5-VL-7B在所有任务中均表现出较高的错误率，尤其在 ViT-G 和 ViT-H 生成的干扰项下最为明显，表明其在应对语义相似干扰项时抗干扰能力较弱。相比之下，GPT-4o和 Gemini-1.5-Pro在 Option domain中表现出更稳定的语义判别能力。

\begin{figure}[htb]
    % \flushleft
    \centering
    \includegraphics[width=1\linewidth]{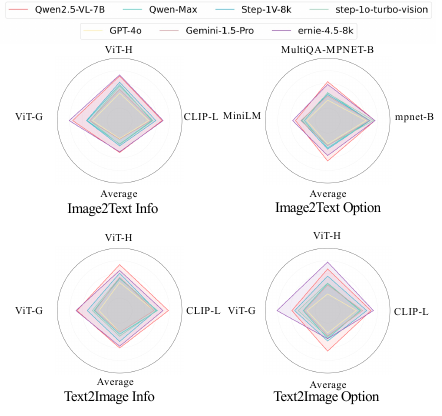}
    \caption{\textbf{We performed a comprehensive analysis of error patterns and their origins across all tested MLLMs. }Models exhibit clear differences in their preferences across distractors selected by different embedding models, revealing variability in how each model responds to semantic interference from different embedding spaces.}
    \label{fig:radar}
\end{figure}

% 相较于 Info domain，Option domain 的错误率普遍更高，说明模型在应对多选干扰时面临更大的挑战；同时，Text2Image 的错误率普遍略高于 Image2Text，说明在“从文本中定位图像”这一任务中，当前模型的科学概念语义抽取与图像匹配能力仍有不足。同时，我们可以发现，text2image任务上不同MLLM偏好的embedding model出现波动的频率会更大，例如在text2image info track上，ernie-4.5-8k更偏好ViT-G筛选的干扰项而Qwen2.5-VL-7B更偏好ViT-H；Gemini-1.5-pro更偏好ViT-H而Step-1o-turbo-vision更偏好ViT-G。这提供了一种方式来研究MLLMs的hidden language：即使用难度更高的text2image模态，通过构建不同形式的干扰项，有效暴露不同模型在跨模态语义理解上的弱点。
\Figref{fig:radar} presents the error rates of several advanced multimodal large language models (MLLMs), including GPT-4o, Gemini-1.5-Pro, Qwen-Max, and the Step series, across four task dimensions. Each model is evaluated against distractors generated using different embedding models (see \secref{para:distractors}). The results reveal clear performance disparities among models across tasks, particularly along the central average axis, indicating that MLLMs exhibit systematic differences in their susceptibility to confusion under various cross-modal matching tasks. Notably, Qwen2.5-VL-7B consistently shows higher error rates across all tasks, with the highest rates observed under distractors generated by ViT-G and ViT-H, suggesting weaker robustness to semantically similar distractors. In contrast, GPT-4o and Gemini-1.5-Pro demonstrate more stable semantic discrimination.

Overall, error rates are higher in the option domain than in the info domain, suggesting that models face greater difficulty when distinguishing between closely related distractors. Additionally, Text2Image tasks yield slightly higher error rates than Image2Text tasks, indicating that current models still struggle with extracting scientific semantic concepts from text and accurately grounding them in visual content. Interestingly, we also observe greater variation in embedding preferences among different MLLMs on the Text2Image tasks. For instance, in the Text2Image info track, ernie-4.5-8k tends to be more confused by ViT-G-generated distractors, while Qwen2.5-VL-7B is more affected by ViT-H. Gemini-1.5-Pro shows greater susceptibility to ViT-H, whereas Step-1o-Turbo-Vision is more challenged by ViT-G. These findings highlight the potential of using more difficult Text2Image settings, along with diverse distractor generation strategies, to effectively uncover weaknesses in the cross-modal semantic alignment capabilities of different MLLMs—thereby offering a lens into the ``hidden language" each model uses to understand visual semantics.
\section{Text-Free \latestBench Analysis}
% 图中展现的是我们在 text-free 的 \latestBench 测试中，GPT-4o 和与论文同期发布的 Gemini 2.5-Pro 都判断错误的一个样例。虽然这两种最先进的 MLLMs 都成功识别出了选项 C 图像中的关键视觉元素——“药片”与“处方单”，展示了它们较强的图像感知（perceptive）能力，但在随后的推理过程中，它们均判断该图像“不包含耐药性相关内容”或“未体现癌症治疗机制”，同时臆断其他选项的细胞类型为癌细胞，最终导致错误。这表明，当前的 MLLMs 虽然能够识别视觉要素本身，但仍缺乏将这些要素与背后深层科学概念（如癌症药物耐药性）建立语义连接的能力。也就是说，它们无法理解“药片 + 处方”在特定科研语境下可能暗示“癌症治疗”或“药物耐受”等科学语义，仍停留在表层元素的拼接识别阶段。这一结果进一步说明，仅靠感知能力尚无法支持模型完成跨模态的科学概念理解任务，必须引入更高层次的推理机制来弥合视觉识别与语义推断之间的鸿沟。
\begin{table}[htb]
\centering
\renewcommand{\arraystretch}{1.5}
\begin{tabular}{ccccc}
\toprule
\textbf{MLLM} & \textbf{Image with Text} & \textbf{Prompt with Text} & \textbf{Acc.(\%)$\uparrow$} & \textbf{ECE}$\uparrow$ \\ \midrule
GPT-4o        & Y                        & N                         & 75.1        & 0.0534       \\
Qwen-VL-Max   & Y                        & N                         & 69.8        & 0.1706       \\
GPT-4o        & N                        & N                         & 64.8        & 0.0419       \\
Qwen-VL-Max   & N                        & N                         & 61.6        & 0.1334       \\
GPT-4o        & N                        & Y                         & 75.5        & 0.0261       \\
Qwen-VL-Max   & N                        & Y                         & 70.8        & 0.2018       \\ \bottomrule
\end{tabular}
\caption{\textbf{Performance Comparison of MLLMs under Different Input Modality Configurations.}We evaluated the performance of two representative models on text-free images of the image2text task under two conditions: with and without supplementing the prompt with text information obtained via OCR. This performance was then compared against the models' baseline results on the unprocessed images. Y denotes `Yes' and N denotes `No'.}
\label{tab:ocr}
\end{table}
\begin{figure}[htb]
    \centering
    \includegraphics[width=0.95\linewidth]{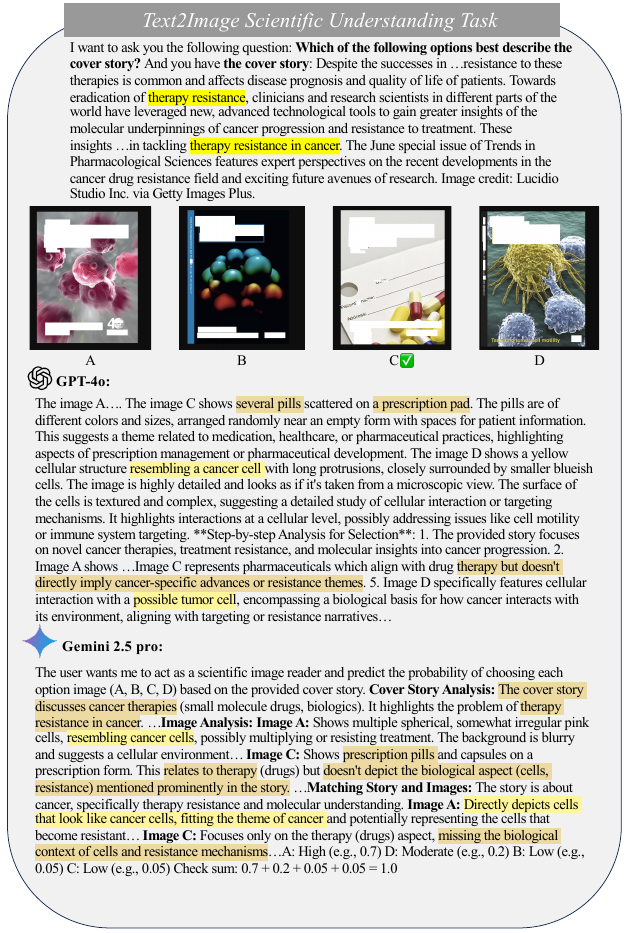}
    \caption{\textbf{Case study of GPT-4o and Gemini 2.5-Pro on the text-free version of \latestBench. }Despite being two of the most advanced MLLMs, both models fail to establish the semantic link between the visual elements and the underlying scientific concept. The \colorbox{pptyellow}{ key scientific concepts from the story} are highlighted in the figure, along with the models’ judgments on both \colorbox{green}{the ground-truth} and \colorbox{bright}{distractor options.}}
    \label{fig:case_study}
\end{figure}
\Figref{fig:case_study} presents a representative failure case from our text-free \latestBench evaluation, where both GPT-4o and the concurrently released Gemini 2.5-Pro~\citep{gemini2.5pro} misidentified the correct answer. While these state-of-the-art MLLMs successfully recognized the key visual elements in option C—namely, ``pills" and a ``prescription pad"—demonstrating strong perceptual capabilities, they subsequently reasoned that the image lacked content related to “drug resistance” or “mechanisms of cancer therapy.” Instead, both models erroneously inferred that other options depicted cancer cells, leading to incorrect choices.

This highlights a fundamental limitation: despite accurately identifying surface-level visual elements, current MLLMs still struggle to establish deeper semantic associations between those elements and the underlying scientific concepts. Specifically, they fail to recognize that in a biomedical context, the combination of “pills + prescription” may imply “cancer treatment” or “drug resistance.” Their understanding remains at the level of visual recognition without fully grasping the latent scientific meaning behind these cues.

This result underscores a broader insight: perceptual ability alone is insufficient for robust multimodal scientific understanding. Bridging the gap between visual recognition and semantic inference requires higher-level reasoning mechanisms that go beyond deriving what it signifies and situating it within broader scientific narratives—toward understanding what it signifies. Not only to recognize, but also to understand—ultimately, to reason.

% This result underscores a broader insight: perceptual ability alone is insufficient for robust multimodal scientific understanding. While recognizing visual elements is a necessary first step, it does not guarantee meaningful comprehension of the underlying scientific concepts. Bridging the gap between visual recognition and semantic inference demands higher-level reasoning mechanisms—ones capable of interpreting abstract relationships, contextual cues, and domain-specific meanings. It requires models not merely to assemble what is seen, but to derive what it signifies, to situate it within broader scientific narratives, and to reason about its implications. Not only to recognize, but also to understand—deeper still, to infer, to explain, and ultimately, to reason.
\section{Models Performance in \ourMethod}
% 表格 5 展示了在我们提出的推理增强框架 \ourMethodShort中，引入不同推理模型对 GPT-4o 性能的影响，用以评估该方法的鲁棒性和通用性，指标包括准确率（Acc.）、置信度校准（ECE）、负对数似然（NLL）以及均方根误差（RMS）。引入任何推理模型后，GPT-4o 在两个任务中准确率均有提升。尤其是 Deepseek-R1 的组合，info domain 的准确率由 0.751 提升至 0.767，option domain 从 0.735 提升至 0.752，均为所有模型中最高，说明引入结构化推理显著提升了 GPT-4o 对视觉科学语义的理解与跨模态推理能力。

% 但不同推理模型带来的提升程度有差异，且可能会导致模型置信度变得不稳定。QwQ-32B 和 GPT-o3-mini-high 同样带来了性能提升，但在 ECE、NLL、RMS 等指标上表现略逊于 Deepseek-R1。QwQ-32B 在 option domain 的 ECE 达到 0.076，高于 Deepseek-R1 的 0.064，说明其预测置信度更不稳定。GPT-o3-mini-high 虽在准确率上有提升，但其 RMS 和 NLL 明显高于其他模型，说明其判断波动更大、信心不够充分。\ourMethodShort表现出与更强大的语言模型搭配（如 Deepseek-R1）时，能进一步挖掘 MLLMs 的科学推理潜力，获得更强的科学理解能力。

\begin{table}[htb]
\centering
\renewcommand{\arraystretch}{1.5}
\adjustbox{max width=\linewidth}{\large
\begin{tabular}{cccccccccc}
\toprule
\multirow{3}{*}{\textbf{MLLM}} & \multirow{3}{*}{\textbf{Reasoning}} & \multicolumn{8}{c}{\textbf{Image2Text Level}}                       \\
                               &                                     & \multicolumn{4}{c}{Info Domain} & \multicolumn{4}{c}{Option Domain} \\
                               &                                     & Acc.(\%)   & ECE    & NLL   & RMS   & Acc.(\%)   & ECE    & NLL    & RMS    \\ \midrule
\multirow{4}{*}{GPT-4o}        & /                                   & 75.1  & \underline{0.053}  & \underline{1.291} & \textbf{0.089} & 73.5  & 0.055  & \underline{1.336}  & \textbf{0.086}  \\
                               & QwQ-32B                             & \underline{76.3}  & 0.059  & 1.766 & 0.092 & 73.6  & 0.076  & 1.429  & 0.099  \\
                               & GPT-o3-mini                         & 75.3  & 0.064  & 1.471 & 0.112 & \underline{74.9}  & \textbf{0.063}  & 1.507  & \underline{0.108}  \\
                               & Deepseek-R1                         & \textbf{76.7}  & \textbf{0.051}  & \textbf{1.243} & \underline{0.092} & \textbf{75.2}  & \underline{0.064}  & \textbf{0.938}  & 0.101  \\ \bottomrule
\end{tabular}
}
\caption{\textbf{Ablation studies of the reasoning model in \ourMethodShort.} We evaluated the robustness of our \ourMethodShort method by substituting its reasoning component with models from various providers and of different parameter scales, consistently demonstrating performance improvements over baseline models.}
\label{tab:reasoning-ablation}
\end{table}

% \begin{table}[htb]
% \centering
% \renewcommand{\arraystretch}{1.5}
% \begin{tabular}{cccccc}
% \toprule
% \textbf{MLLM}  & \textbf{Close Source} & \textbf{MMBench V1.1} & \textbf{MathVista} & \textbf{AI2D} & \textbf{MMVet} \\ \midrule
% Step-1o        & Y                     & 87.3                  & 74.7               & 89.1          & 82.8           \\
% SenseNova      & Y                     & 85.7                  & 78.4               & 87.8          & 78.2           \\
% Gemini1.5-pro  & Y                     & 82.8                  & 67.9               & 83.3          & 74.6           \\
% GLM-4v-Plus    & Y                     & 85.9                  & 73.5               & 86.7          & 75.7           \\
% Qwen2.5-VL-72B & N                     & 87.8                  & 74.2               & 88.5          & 76.9           \\
% GPT-4o         & Y                     & 83.4                  & 70.5               & 87.2          & 75.3           \\ \bottomrule
% \end{tabular}
% \end{table}
\Tabref{tab:reasoning-ablation} presents the performance impact of incorporating different reasoning models into our proposed reasoning-augmented framework, \ourMethodShort, using GPT-4o as the base model. We evaluate accuracy (Acc.), expected calibration error (ECE), negative log-likelihood (NLL), and root mean square error (RMS) to assess robustness and generalizability. Across both tasks, introducing any reasoning model leads to improved accuracy for GPT-4o. Notably, the combination with Deepseek-R1 yields the highest performance, with accuracy in the info domain increasing from 0.751 to 0.767 and in the option domain from 0.735 to 0.752. These results highlight the effectiveness of structured reasoning in enhancing GPT-4o’s scientific semantic understanding and cross-modal reasoning capabilities.

However, the degree of improvement varies across reasoning models, and some may introduce confidence instability. While QwQ-32B and GPT-o3-mini-high also boost accuracy, their performance in ECE, NLL, and RMS is slightly inferior to that of Deepseek-R1. For instance, QwQ-32B shows a higher ECE of 0.076 in the option domain compared to Deepseek-R1’s 0.064, suggesting less stable confidence calibration. Although GPT-o3-mini-high improves accuracy, it exhibits noticeably higher RMS and NLL, indicating greater prediction variance and weaker confidence. These findings suggest that \ourMethodShort, when paired with more advanced reasoning models such as Deepseek-R1, can further unlock the scientific reasoning potential of MLLMs and lead to more robust multimodal understanding.

\section{Performance of \textit{Live} Data Curation}
\begin{table}[htb]
\centering
\renewcommand{\arraystretch}{1.5}
\begin{tabular}{cccccccc}
\toprule
\multirow{2}{*}{\textbf{MLLMs}} & \multirow{2}{*}{\textbf{Dataset Construction}} & \multicolumn{4}{c}{\textbf{Image2Text Level}} \\
                                &                                                & Acc(\%)$\uparrow$.      & ECE$\downarrow$       & NLL$\downarrow$       & RMS$\downarrow$       \\ \midrule
\multirow{2}{*}{Qwen2.5-VL-7B}  & 2023 Embeddings                                   & 59.7     & \textbf{0.055}     & \textbf{1.856}     & \textbf{0.095}     \\
                                & 2025 Embeddings                                 & \textbf{58.4}     & \underline{0.061}     & \underline{1.923}     & \underline{0.105}     \\
\multirow{2}{*}{Qwen-VL-Max}       & 2023 Embeddings                                  & 69.8     & 0.171     & 3.054     & 0.210     \\
                                & 2025 Embeddings                                  & 68.7     & 0.205     & 3.472     & 0.240     \\
\multirow{2}{*}{Gemini-1.5-Pro} & 2023 Embeddings                                  & \textbf{72.7}     & 0.108     & 5.058     & 0.153     \\
                                & 2025 Embeddings                                  & 71.8     & 0.128     & 5.567     & 0.165 \\ \bottomrule
\end{tabular}
\caption{\textbf{MLLMs' Performance on Benchmarks Generated Using State-of-the-Art Embedding Models. }We reconstructed new distractor items for samples identical to those in MAC-2025 from the MAC dataset using three state-of-the-art embedding models, and evaluated baseline models' performance on the image2text task.}
\label{tab:mac2026}
\end{table}
% 为了详细探究\emph{live} data curation的效果，我们基于2025年发布的三种先进嵌入模型，重新构建了\latestBench中2287个样本的干扰项得到2025embedding版本的评测集。我们选取了三种代表性MLLMs（Qwen2.5-VL-7B、Qwen-Max 和 Gemini-1.5-Pro），并分别对比其在使用2023年嵌入模型与2025年嵌入模型在\latestBench上构建的基准的表现差异。

% 实验结果显示，使用2025Embeddings构建的\latestBench上均表现出了模型准确率的下降，Qwen2.5-VL-7B下降了1.3%，Qwen-Max下降了1.1%，Gemini-1.5-Pro下降了0.9%。不仅如此，所有模型均出现了NLL与RMS显著升高，表明模型在该版本上的预测不确定性更强，置信度更不稳定。此外，所有模型在ECE指标上也普遍升高，例如Qwen-Max的ECE从0.171升至0.205，说明新一代嵌入模型能够生成更加语义接近、难以区分的干扰项，从而显著提高了任务的混淆度与挑战性。

% 这一结果进一步印证了我们“live benchmark”设计理念的必要性与有效性：随着嵌入模型和表示学习技术的不断演进，定期引入更强大的嵌入模型生成干扰项，能够持续提升基准的前沿性与挑战性，从而更真实、更精准地衡量当代 MLLM 在科学语义建模与跨模态理解中的能力上限。

To further investigate the effectiveness of our \emph{live} data curation strategy, we reconstruct the distractors for all 2,287 samples in \latestBench using three advanced embedding models released in 2025, resulting in the 2025-embedding version of the benchmark. We select three representative MLLMs—Qwen2.5-VL-7B, Qwen-Max, and Gemini-1.5-Pro—and compare their performance on benchmarks constructed with both 2023 and 2025 embedding models.

The results in \tabref{tab:mac2026} show a consistent drop in accuracy across all models when evaluated on the 2025 embedding version: Qwen2.5-VL-7B drops by 1.3\%, Qwen-Max by 1.1\%, and Gemini-1.5-Pro by 0.9\%. All models exhibit a noticeable increase in NLL and RMS, indicating greater prediction uncertainty and reduced confidence stability. ECE values also increase across the board—for example, Qwen-Max’s ECE rises from 0.171 to 0.205—suggesting that the newer embedding models generate more semantically similar and harder-to-distinguish distractors, thereby raising the task’s difficulty to \say{former} MLLM.

These findings further validate the necessity and effectiveness of our \emph{live} benchmark design. As embedding models and representation learning techniques continue to evolve, periodically incorporating stronger embedding models to generate distractors helps maintain the benchmark’s cutting-edge nature, enabling more accurate assessment of the limits of contemporary MLLMs in scientific semantic modeling and cross-modal understanding.

\end{document}